\documentclass{article} 
\pdfoutput=1
\usepackage{iclr2026_conference,times}

\usepackage{amsmath,amsfonts,bm}

\def\eqref#1{equation~\ref{#1}}

\def\1{\bm{1}}

\DeclareMathAlphabet{\mathsfit}{\encodingdefault}{\sfdefault}{m}{sl}
\SetMathAlphabet{\mathsfit}{bold}{\encodingdefault}{\sfdefault}{bx}{n}

\usepackage{verbatim}
\usepackage[cache]{minted}
\usepackage{xcolor}
\usepackage{listings}
\usepackage{algorithm}

\definecolor{keywordcolor}{RGB}{0,0,255}
\definecolor{stringcolor}{RGB}{163,21,21}
\definecolor{commentcolor}{RGB}{0,128,0}

\usepackage[utf8]{inputenc} 
\usepackage[T1]{fontenc}    
\usepackage{xcolor}         
\usepackage{subcaption}
\definecolor{linkColor}{rgb}{0.2,0.4,0.6}
\usepackage[colorlinks=true,linkcolor=linkColor,citecolor=linkColor,filecolor=linkColor,urlcolor=linkColor]{hyperref}
\usepackage{hyperref}       
\usepackage{url}            
\usepackage{booktabs}       
\usepackage{amsfonts}       
\usepackage{nicefrac}       
\usepackage{microtype}      
\usepackage{xcolor}         
\usepackage{times}
\usepackage{amsmath}
\usepackage{amssymb}
\usepackage{graphicx}
\usepackage{etoc}
\usepackage{multirow}
\usepackage{multicol}
\usepackage{algorithmic}
\usepackage{booktabs}
\usepackage{enumitem}
\usepackage{mathrsfs}
\usepackage{colortbl}  
\usepackage{array}   
\usepackage{pifont}
\usepackage{xspace}
\definecolor{gray0}{rgb}{0.8,0.9,1.0}

\newcommand{\Ours}{\textup{GMPO}\xspace}

\usepackage{bm}
\usepackage{listings}
\usepackage{makecell}  
\usepackage{diagbox} 
\usepackage{mathtools} 
\usepackage{graphicx} 
\usepackage{multirow}
\usepackage{wrapfig}

\usepackage{hyperref}
\usepackage{url}

\title{Geometric-Mean Policy Optimization}

\author{Yuzhong Zhao\textsuperscript{1}\thanks{~Equal contribution. Work done during internship at Microsoft Research.}  \quad 
Yue Liu\textsuperscript{1}\footnotemark[1]  \quad
Junpeng Liu\textsuperscript{2}\quad
Jingye Chen\textsuperscript{3}\quad
Xun Wu\textsuperscript{4}\quad
Yaru Hao\textsuperscript{4}\\
\textbf{Tengchao Lv}\textsuperscript{4}\quad
\textbf{Shaohan Huang}\textsuperscript{4}\quad
\textbf{Lei Cui}\textsuperscript{4}\quad
\textbf{Qixiang Ye}\textsuperscript{1}\quad
\textbf{Fang Wan}\textsuperscript{1}\quad
\textbf{Furu Wei}\textsuperscript{4}\\
\textsuperscript{1}UCAS
\quad
\textsuperscript{2}CUHK
\quad
\textsuperscript{3}HKUST
\quad
\textsuperscript{4}Microsoft Research \\
{\href{https://aka.ms/GeneralAI}{https://aka.ms/GeneralAI}}
}

\iclrfinalcopy 
\begin{document}

\maketitle

\begin{abstract}
Group Relative Policy Optimization (GRPO) has significantly enhanced the reasoning capability of large language models by optimizing the arithmetic mean of token-level rewards.
Unfortunately, GRPO is observed to {suffer from unstable policy updates} when facing tokens with outlier importance-weighted rewards, which manifest as extreme importance sampling ratios during training.
In this study, we propose \textit{\textbf{G}eometric-\textbf{M}ean \textbf{P}olicy \textbf{O}ptimization} (\textbf{\Ours}), with the aim to improve the stability of GRPO through {suppressing token reward outliers}.
GMPO is plug-and-play—simply replacing GRPO's arithmetic mean with the geometric mean of token-level rewards, as the latter is inherently less sensitive to outliers.
GMPO is theoretically plausible—analysis reveals that both GMPO and GRPO are weighted forms of the policy gradient while the former enjoys more stable weights, which consequently benefits policy optimization and performance.
Experiments on multiple mathematical reasoning benchmarks show that \Ours-7B improves the average Pass@1 of GRPO by up to 4.1\%, outperforming many state-of-the-art approaches.
Code is available at 
{\href{https://github.com/callsys/GMPO}{https://github.com/callsys/GMPO}}.

\begin{figure*}[h]
	\includegraphics[width=1.0\linewidth]{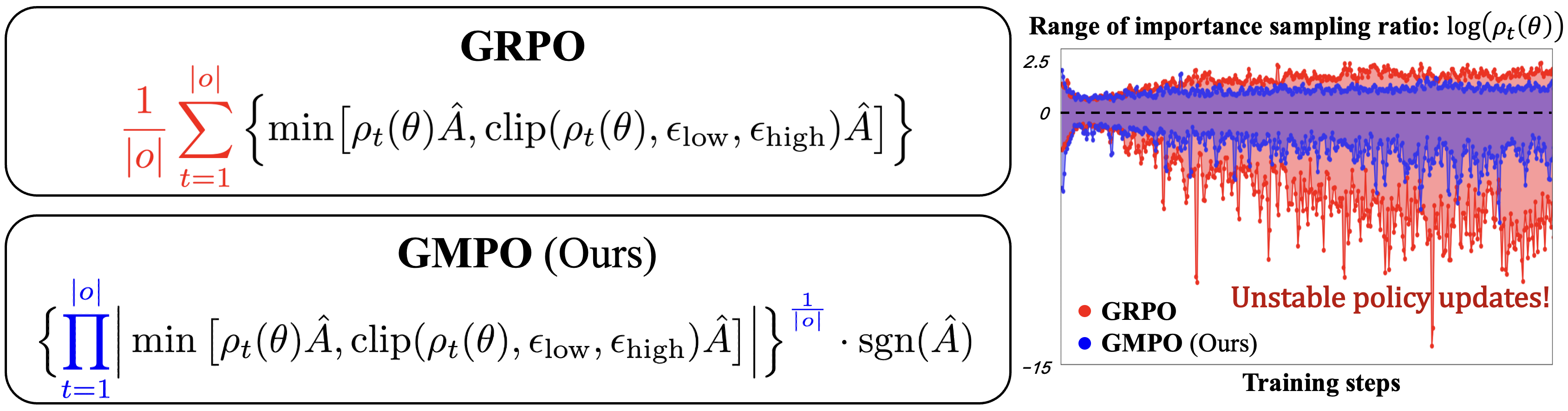}
     \caption{Comparison between GRPO and our \Ours. GRPO optimizes the arithmetic mean of token-level rewards while \Ours the geometric mean (left). 
     When training with GRPO, the important sample ratio $\big(\rho_{t}(\theta)=\frac{\pi_{\theta}(o_{t}|q,o_{<t})}{\pi_{\theta_{\mathrm{old}}}(o_{t}|q,o_{<t})}\big)$ frequently reaches extreme values, leading to unstable policy updates. In contrast, \Ours enjoys more stable important sample ratio with fewer outliers (right).
     }
    \label{fig:1}
\end{figure*}

\end{abstract}

\section{Introduction}
\label{sec:intro}

As test-time scaling becomes a key research focus in the large language model community, recent post-training methods have increasingly sought to extend chain-of-thought (CoT) generation to enhance reasoning capabilities.
Recent advances, such as Group Relative Policy Optimization (GRPO)~\citep{shao2024deepseekmath}, leverage multiple sampled responses per input prompt to compute relative rewards and advantages ($\hat{A}$ in Figure\ref{fig:1}, left), leading to notable improvements in reasoning performance.
By maximizing the arithmetic mean of token-level rewards, GRPO has achieved strong results on complex tasks such as mathematics, code generation, and multimodal reasoning.

During GRPO training, the importance-weighted reward for each token is given by $\rho_{t}(\theta) \hat{A}$, where the important sampling ratio $\rho_{t}(\theta)$ is defined as $\rho_{t}(\theta)=\frac{\pi_{\theta}(o_{t}|q,o_{<t})}{\pi_{\theta_{\mathrm{old}}}(o_{t}|q,o_{<t})}$. 
This ratio plays a key role in PPO~\citep{schulman2017proximal} and GRPO, ensuring that policy updates are grounded in data from the current policy $\pi_{\theta}$.
Large deviations of $\rho_{t}(\theta)$ from 1 indicate excessive policy shifts, leading to overly aggressive updates and instability. 
Constraining the ratio within a reasonable range is therefore critical for stable and reliable training.

\begin{figure*}[t]
	\includegraphics[width=1.0\linewidth]{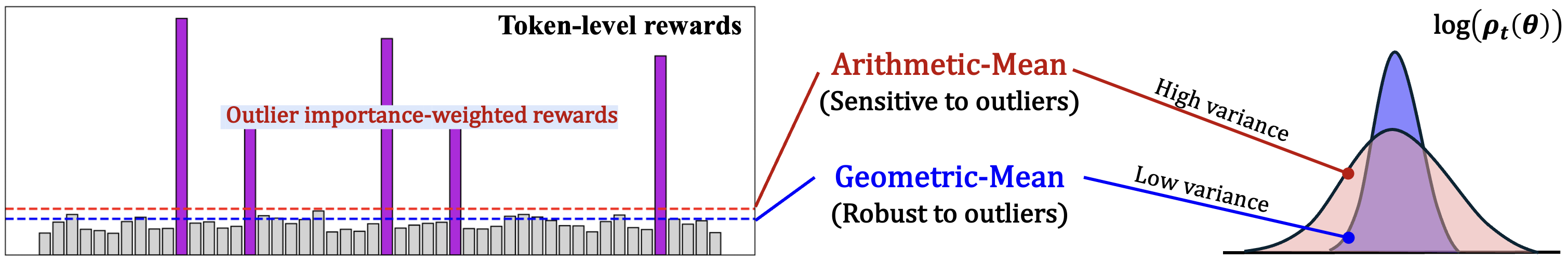}
     \caption{Compared to the arithmetic mean, the geometric mean is more robust to outliers and yields importance sampling ratio distributions with lower variance.
     }
    \label{fig:2}
\end{figure*}

As shown in Figure~\ref{fig:1} (top left), objective of GRPO involves the arithmetic mean of token-level rewards, which is sensitive to outliers (Figure~\ref{fig:2}). 
As training progresses (Figure~\ref{fig:1}, right), the range of $\rho_{t}(\theta)$ under GRPO expands, leading to unstable policy updates and degraded model performance.
To mitigate this, GRPO applies a clipping range $(\epsilon_{\mathrm{low}}, \epsilon_{\mathrm{high}})$ to restrict large deviations of $\rho_{t}(\theta)$. 
However, this constraint causes limited exploration and early deterministic policy,
which can hinder the scaling process~\citep{yu2025dapo}.

To alleviate the instability while enhancing exploration capabilities of GRPO, we propose \textit{\textbf{G}eometric-\textbf{M}ean \textbf{P}olicy \textbf{O}ptimization} (\textbf{\Ours}), Figure~\ref{fig:1} (left bottom). 
%
{\Ours takes full advantages of} the geometric mean, which is inherently less sensitive to outliers and yields
importance sampling ratio distributions with lower variance (Figure~\ref{fig:2}).
During training (Figure~\ref{fig:1}, right), the range of \Ours’s $\rho_{t}(\theta)$ remains stable, exhibiting fewer extreme values than GRPO. 
With GMPO, we can maintain stable policy optimization while allowing a larger clipping range to promote greater exploration.

To further emphasize the advantages of \Ours, we provide detailed theoretical and experimental analyses to justify its training objective. 
First, we show that \Ours’s objective produces a narrower value range than GRPO’s, indicating reduced training variance and more stable policy updates.
Second, from a gradient perspective, \Ours provides a more balanced update signal and is more robust to outlier values of the importance sampling ratio $\rho_{t}(\theta)$.
Third, as training progresses, \Ours maintains a smaller KL divergence from the pre-trained model and higher token entropy than GRPO, indicating enhanced stability (via smaller KL) and greater policy exploration (via higher entropy).

Extensive experiments on both language and multimodal reasoning tasks demonstrate the advantages of \Ours over GRPO. Specifically, on five mathematical reasoning benchmarks of varying difficulty (AIME24~\citep{li2024numinamath}, AMC~\citep{li2024numinamath}, MATH500~\citep{hendrycks2021measuring}, Minerva~\citep{lewkowycz2022solving}, and OlympiadBench~\citep{olympiadbench}), \Ours improves the average Pass@1 accuracy by 4.1\% (63.4\% vs. 59.3\%) with DeepSeek-R1-Distill-Qwen-7B compared to GRPO.
Besides, \Ours improves the Pass@1 accuracy by 2.1\% (96.7\% vs. 94.6\%) on MATH500 with a Qwen-32B~\citep{yang2025qwen3} Mixture-of-Experts model.
On Geometry3K multimodal reasoning benchmark~\citep{geometry3k}, \Ours increases the average Pass@1 accuracy by 1.4\% (54.7\% vs. 53.3\%) with Qwen2.5-VL-Instruct-7B.

The contributions of this study are summarized as follows:
\begin{itemize}[leftmargin=*]
    \item We propose \textit{\textbf{G}eometric-\textbf{M}ean \textbf{P}olicy \textbf{O}ptimization} (\textbf{\Ours}), which stabilizes the GRPO algorithm by maximizing the geometric mean of token-level rewards.

    \item We conduct thorough theoretical and empirical analyses, showing that \Ours improves stability while enhancing exploration relative to GRPO.

    \item \Ours-7B achieves 4.1\% higher Pass@1 accuracy than GRPO-7B on five mathematical reasoning benchmarks, and 1.4\% higher accuracy on the Geometry3K multimodal reasoning benchmark.
    
\end{itemize}
\section{Background}
\subsection{Related Works}
\label{sec:relate_work}
Reinforcement learning (RL) has become a key approach for post-training large language models (LLMs), with verifiable rewards enabling significant reasoning improvements, as demonstrated by DeepSeek-R1~\citep{deepseekr1}. Building on Proximal Policy Optimization (PPO) \citep{schulman2017proximal}, numerous variants have been developed to enhance efficiency and performance.

GRPO \citep{shao2024deepseekmath,deepseekr1} eliminates the need for computationally expensive value models while maintaining strong results across mathematics, coding, and QA benchmarks. GPG \citep{chu2025gpg} further simplifies optimization by eliminating surrogate losses, critics, and KL constraints. Several extensions address rollout selection or bias correction: SRPO \citep{zhang2025srpo} uses history resampling, DAPO \citep{yu2025dapo} employs dynamic sampling, Dr.GRPO \citep{drgrpo} mitigates length bias, and OPO \citep{opo} provides an optimal baseline to reduce gradient variance. Reward shaping and advantage estimation are also actively explored. EMPO \citep{zhang2025empo} incorporates semantic entropy, AAPO \citep{xiong2025aapo} introduces advantage momentum, and BNPO \citep{xiao2025bnpo} adaptively normalizes rewards via a Beta distribution. Seed-GRPO \citep{seed_grpo} scales policy updates by question uncertainty, while GRPO-lead \citep{grpo_lead} addresses reward sparsity through length-dependent accuracy, explicit penalties, and difficulty-aware reweighting. Efficiency-driven methods include CPPO \citep{lin2025cppo} (pruning low-advantage completions), S-GRPO \citep{sgrpo} (early exit to cut redundancy), Ada-GRPO \citep{wu2025arm} (adaptive reasoning formats), and GVPO \citep{gvpo} (analytical KL-constrained weighting). GRPO-$\lambda$ \citep{dai2025stablerl} dynamically switches between length-penalized and length-agnostic rewards to avoid collapse. Further methods improve rollout usage. PODS \citep{xu2025not_all_rollouts} trains only on informative subsets of parallel rollouts, while RePO \citep{li2025repo} retrieves diverse off-policy samples via replay. RAFT \citep{xiong2025minimalist} trains solely on positive samples yet rivals GRPO. INTUITOR \citep{zhao2025learning_external_rewards} eliminates external rewards by using model self-certainty, and PRIME \citep{prime} provides a scalable RL framework for reasoning.
Exploration-focused techniques include the 80/20 rule \citep{8020rule}, which emphasizes high-entropy minority tokens, and entropy-based advantage augmentation \citep{entropy_perspective}.
Complementary to algorithmic advances, data-centric approaches have also proven crucial. Open-Reasoner-Zero \citep{open_reasoner} curates 129k diverse, high-quality samples with curriculum learning. Eurus \citep{eurus} contributes a large-scale alignment dataset and novel reward modeling. 

Despite rapid progress, the stability of RL for LLMs remains underexplored, yet it is essential for developing reliable and scalable post-training systems.

\subsection{Preliminary}
\label{sec:preliminary}

The Group Relative Policy Optimization algorithm is initially proposed in DeepSeek-math~\citep{shao2024deepseekmath}. The core idea is to estimate the baseline through a relative reward within a group of rollouts, which reduces the computational cost of the critic model and improves the training stability. Specifically, for each question $q$ from the training set $Q$, GRPO samples a group of rollouts $\{o_1, o_2, \cdots, o_G\}$ from the old policy $\pi_{\theta_\mathrm{old}}$ and calculates the corresponding rewards $\{r_1, r_2, \cdots, r_G\}$. Then the policy model $\pi_{\theta}$ is optimized by maximizing the following objective:
{
\begin{align}
\mathcal{J}_{\mathrm{GRPO}}(\pi_{\theta}) &= \mathbb{E}_{q\sim \mathcal{Q}, {\{o_i\}_{i=1}^G\sim \pi_{\theta_{\mathrm{old}}}(\cdot|q)}}\notag\\
\frac{1}{G}&\sum_{i=1}^{G}\frac{1}{|o_i|}\sum_{t=1}^{|o_i|} \left\{\min \big [\rho_{i,t}(\theta)\hat{A}_{i}, \mathrm{clip}(\rho_{i,t}(\theta), \epsilon_\mathrm{low}, \epsilon_\mathrm{high}) \hat{A}_{i}\big ]-\beta \mathrm{D}_{\mathrm{KL}}(\pi_\theta\parallel\pi_{\mathrm{ref}})\right\}, \label{eq:grpo}
\end{align}
}
where $\rho_{i,t}(\theta)=\frac{\pi_{\theta}(o_{i,t}|q,o_{i,<t})}{\pi_{\theta_{\mathrm{old}}}(o_{i,t}|q,o_{i,<t})}$ and $\hat{A}_i = \frac{r_i - \mathrm{mean}(\{r_1, r_2, \cdots r_G\})}{\mathrm{std}(\{r_1, r_2, \cdots r_G\})}$. $\rho_{i,t}(\theta)$ represents the importance sampling ratio of the $t$-th token in the $i$-th rollout based on the current policy $\pi_{\theta}$ and old policy $\pi_{\theta_\mathrm{old}}$. 
$\hat{A}_i$ is the advantage of the $i$-th rollout and is calculated by normailizing the rewards that belong to the same group according to GRPO. $(\epsilon_\mathrm{low}, \epsilon_\mathrm{high})$ are the clipping thresholds and $\mathrm{D}_{\mathrm{KL}}(\pi_\theta\parallel\pi_{\mathrm{ref}})$ is the KL regularization term.
Following Dr. GRPO~\citep{drgrpo}, we ignore $\mathrm{D}_{\mathrm{KL}}(\pi_\theta\parallel\pi_{\mathrm{ref}})$ for simplicity and memory saving.
The objective of GRPO is equivalent to the arithmetic mean of token-level rewards (We ignore the clipping range term for clarity), which can be formatted as:
{
\begin{align}
\mathcal{J}^*_{\mathrm{GRPO}}(\pi_{\theta}) &= \mathbb{E}_{q\sim \mathcal{Q}, {\{o_i\}_{i=1}^G\sim \pi_{\theta_{\mathrm{old}}}(\cdot|q)}} \left[\frac{1}{G}\sum_{i=1}^G\frac{1}{|o_i|}\sum_{t=1}^{|o_i|}\rho_{i,t}(\theta)\hat{A}_i\right] \label{eq:grpo_reward}.
\end{align}
}

In practice, the rollouts are sampled from the old policy $\pi_{\theta_{\mathrm{old}}}$. To approximate policy updates as if they were based on rollouts sampled from the current policy $\pi_{\theta}$, the normalized advantage $\hat{A}_i$ of each rollout is weighted by the importance sampling ratio $\rho_{i,t}(\theta)$.

\section{Geometric-Mean Policy Optimization}
\label{sec:method}

As shown in Figure~\ref{fig:1}(right), we observe tokens with extreme importance sampling ratios during GRPO training, indicating unreliable model updates.
This instability arises because GRPO's objective is sensitive to outlier values of importance-weighted rewards, which drive aggressive policy updates and further amplify the variance of importance sampling ratios.

To solve that, we propose \textit{\textbf{G}eometric-\textbf{M}ean \textbf{P}olicy \textbf{O}ptimization} (\textbf{\Ours}), a stabilized variant of GRPO. Instead of optimizing the arithmetic mean of token-level rewards as shown in Equation~\ref{eq:grpo_reward}, \Ours maximizes the geometric mean of them:
\begin{align}
\mathcal{J}^*_{\mathrm{GMPO}}(\pi_{\theta}) &= \mathbb{E}_{q\sim \mathcal{Q}, {\{o_i\}_{i=1}^G\sim \pi_{\theta_{\mathrm{old}}}(\cdot|q)}} \left[\frac{1}{G}\sum_{i=1}^G\Big(\prod_{t=1}^{|o_i|}\big|\rho_{i,t}(\theta)\hat{A}_i\big|\Big)^{\frac{1}{|o_i|}}\cdot\mathrm{sgn}(\hat{A}_i)\right] \label{eq:gmpo_reward},
\end{align}
where $\mathrm{sgn}(\hat{A}_i)$ ensures the correct optimization direction, returning 1 when $\hat{A}_i$ is positive and -1 otherwise. $\mathcal{J}^*_{\mathrm{GMPO}}(\pi_{\theta})$ has a narrower value range than $\mathcal{J}^*_{\mathrm{GRPO}}(\pi_{\theta})$, which can be derived as: 
\begin{align}
|\mathcal{J}^*_{\mathrm{GMPO}}(\pi_{\theta})| &= \mathbb{E}_{q\sim \mathcal{Q}, {\{o_i\}_{i=1}^G\sim \pi_{\theta_{\mathrm{old}}}(\cdot|q)}} \left[\frac{1}{G}\sum_{i=1}^G\Big(\prod_{t=1}^{|o_i|}\big|\rho_{i,t}(\theta)\hat{A}_i\big|\Big)^{\frac{1}{|o_i|}}\right]\notag \\
&\leq \mathbb{E}_{q\sim \mathcal{Q}, {\{o_i\}_{i=1}^G\sim \pi_{\theta_{\mathrm{old}}}(\cdot|q)}} \left[\frac{1}{G}\sum_{i=1}^G\frac{1}{|o_i|}\sum_{t=1}^{|o_i|}\big|\rho_{i,t}(\theta)\hat{A}_i\big|\right]\notag = |\mathcal{J}^*_{\mathrm{GRPO}}(\pi_{\theta})|.\notag
\end{align}
This narrower range suggests that the training process of \Ours experiences lower variance in the optimization objective, which can be viewed as evidence of more stable policy updates.
Compared to $\mathcal{J}_{\mathrm{GRPO}}(\pi_{\theta})$, $\mathcal{J}_{\mathrm{GMPO}}(\pi_{\theta})$ is less sensitive to outliers because the geometric mean is inherently more robust to outliers than the arithmetic mean. As a result, $\mathcal{J}_{\mathrm{GMPO}}(\pi_{\theta})$ provides more reliable policy updates and maintains a more stable range of importance sampling ratios as shown in Figure~\ref{fig:1}(right). 
By expanding Equation~\ref{eq:gmpo_reward} and incorporating the clipping range term from PPO~\citep{schulman2017proximal} at the token-level, we can derive the complete objective function of \Ours as follows:
\begin{align}
\mathcal{J}_{\mathrm{GMPO}}(\pi_{\theta}) &= \mathbb{E}_{q\sim \mathcal{Q}, \{o_i\}_{i=1}^G \sim \pi_{\theta_{\mathrm{old}}}(\cdot|q)}\notag\\
\frac{1}{G}&\sum_{i=1}^{G}\left\{\prod_{t=1}^{|o_i|}\Big|\min \big [\rho_{i,t}(\theta)\hat{A}_{i}, \mathrm{clip}(\rho_{i,t}(\theta), \epsilon_\mathrm{low}, \epsilon_\mathrm{high}) \hat{A}_{i}\big ]\Big|\right\}^{\frac{1}{|o_i|}}\cdot\mathrm{sgn}(\hat{A}_{i}).\label{eq:gmpo}
\end{align}
GMPO is straightforward to implement, and its pseudo-code is given in Algorithm~\ref{alg:gmpo}. For numerical stability, both the product and clipping operations in Equation~\ref{eq:gmpo} are carried out in log space.

To better understand why \Ours is more stable than GRPO, we show that \Ours is more robust to tokens with extreme importance sampling ratios from a gradient perspective. Specifically, given question $q$ and rollout $o_i$, the gradients of $\mathcal{J}^*_{\mathrm{GRPO}}(\pi_{\theta})$ (Equation~\ref{eq:grpo_reward}) and $\mathcal{J}^*_{\mathrm{GMPO}}(\pi_{\theta})$ (Equation~\ref{eq:gmpo_reward}) with respect to the model parameter $\theta$ are as follows\footnote{Clipping range term is omitted for clarity. Detailed derivations are provided in Appendix~\ref{app:derivation}}:
\begin{align}
\scalebox{1.2}{$\displaystyle \nabla$}_\theta\mathcal{J}^*_{\mathrm{GRPO}}(\pi_\theta)\Big|_{q, o_i} &= \frac{1}{G\cdot|o_i|}\sum_{t=1}^{|o_i|}{\color[rgb]{1.0,0,0}\rho_{i,t}(\theta)}\cdot \hat{A}_i\cdot\scalebox{1.2}{$\displaystyle \nabla$}_\theta \mathrm{log}(\pi_{\theta}(o_{i,t}|q,o_{i,<t})),\\
\scalebox{1.2}{$\displaystyle \nabla$}_\theta\mathcal{J}^*_{\mathrm{GMPO}}(\pi_\theta)\Big|_{q, o_i} &= \frac{1}{G\cdot|o_i|}\sum_{t=1}^{|o_i|}{\color[rgb]{0,0,1.0}\Big(\prod_{k=1}^{|o_i|}\rho_{i,k}(\theta)\Big)^{\frac{1}{|o_i|}}}\cdot \hat{A}_i\cdot\scalebox{1.2}{$\displaystyle \nabla$}_\theta \mathrm{log}(\pi_{\theta}(o_{i,t}|q,o_{i,<t})),
\end{align}

The term $\hat{A}_i\cdot\scalebox{1.2}{$\displaystyle \nabla$}_\theta \mathrm{log}(\pi_{\theta}(o_{i,t}|q,o_{i,<t}))$ quantifies the influence of the generated token $o_{i,t}$ on the parameters $\theta$, which corresponds to the standard policy gradient~\citep{policy_gradient}. The gradients of both objectives are weighted sums of the policy gradients of the generated tokens, but with different weights. For $\mathcal{J}^*_{\mathrm{GRPO}}(\pi_{\theta})$, the weight of the token $o_{i,t}$ includes its individual importance sampling ratio $\rho_{i,t}(\theta)$. An extreme $\rho_{i,t}(\theta)$ will cause the token gradient to be too large or small, resulting in aggressive policy updates. For $\mathcal{J}^*_{\mathrm{GMPO}}(\pi_{\theta})$, the weight of the token $o_{i,t}$ includes the geometric mean of all the ratios $\left(\prod_{k=1}^{|o_i|}\rho_{i,k}(\theta)\right)^{{\frac{1}{|o_i|}}}$ in the same sequence, provides a more balanced update signal and is more robust to outlier values. 


\begin{algorithm}[t]
\scriptsize
\caption{\Ours Training Objective}
\label{alg:gmpo}
\begin{lstlisting}
def gmpo_loss(new_probs, old_probs, mask, advantage, epsilon=0.4):
    """
    new_probs [L, 1]: Token probabilities from the current model
    old_probs [L, 1]: Token probabilities from the old model
    mask [L, 1]: Indicates valid (non-padded) tokens
    advantage [1]: Advantage or normalized reward for the sequence
    epsilon [1]: Controls the clipping range
    """
    # Clipping at token-level & Clipping wider
    new_log_probs, old_log_probs = torch.log(new_probs), torch.log(old_probs)
    sgn_A = 1 if advantage > 0 else -1
    sgn_A_log_probs_diff = sgn_A * (new_log_probs - old_log_probs)
    sgn_A_log_probs_diff2 = torch.clamp(sgn_A_log_probs_diff, -epsilon, epsilon)
    sgn_A_log_probs_diff_min = torch.min(sgn_A_log_probs_diff, sgn_A_log_probs_diff2)
    log_probs_diff_min = sgn_A * sgn_A_log_probs_diff_min
    # Geometric-Mean Policy Optimization
    importance_sampling_ratio = torch.exp(log_probs_diff_min[mask].sum()/mask.sum())
    loss = -advantage * importance_sampling_ratio
    return loss
\end{lstlisting}
\end{algorithm}

\begin{figure}
    \centering
    \includegraphics[width=1.0\linewidth]{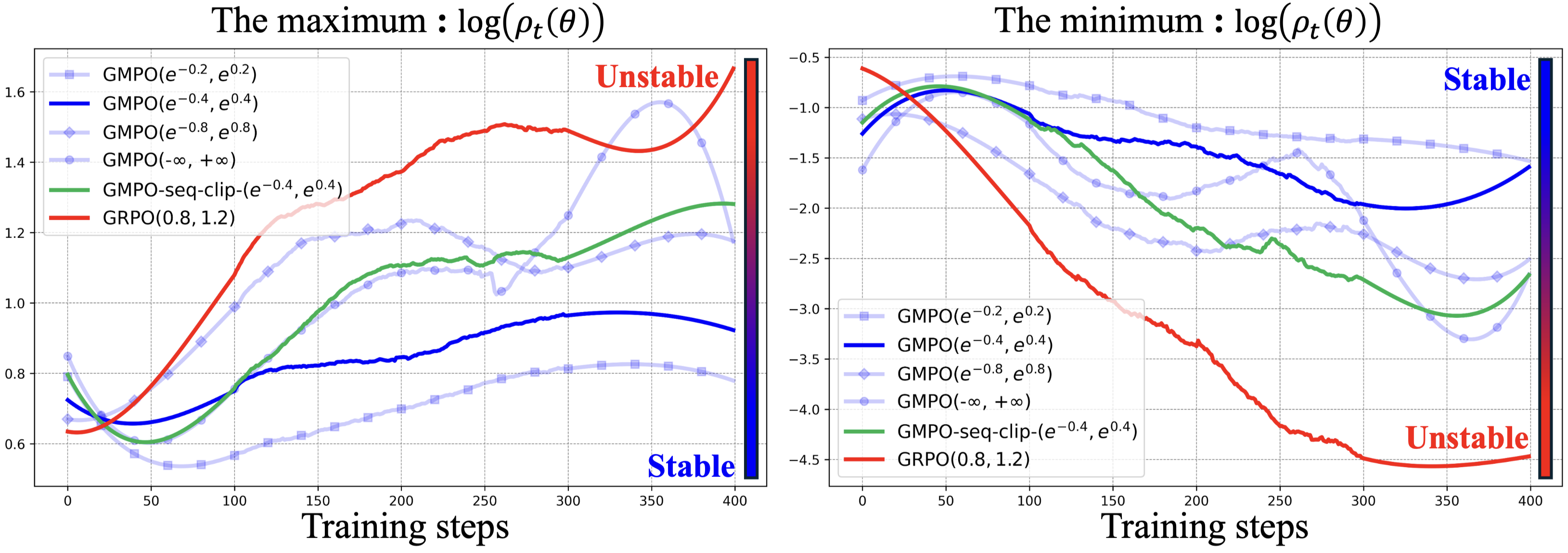}
    \caption{The range of importance sampling ratio $\rho_t(\theta)$ with respect to different clipping range and training steps. A wider range indicates less stable policy updates. Compared to GRPO with a clipping range of (0.8, 1.2), \Ours demonstrates greater stability, even with a larger clipping range of ($e^{-0.4}, e^{0.4}$). All curves are smoothed for clarity.}
    \label{fig:ratio_with_different_clip}
\end{figure}
Beyond the proposed training objective, we demonstrate the effectiveness of the following key designs in \Ours:

\noindent\textbf{(i) Clipping at token-level.} 
Unlike vanilla GRPO in DeepSeek-math~\citep{shao2024deepseekmath}, 
DeepSeek-R1~\citep{deepseekr1} maximizes the sequence-level reward $(\prod_{t=1}^{|o_i|}\rho_{i,t}(\theta))\hat{A}_i$ and clips outliers at the sequence-level, $i.e.$, $\mathrm{clip}\big(\prod_{t=1}^{|o_i|}\rho_{i,t}(\theta),\epsilon_\mathrm{low},\epsilon_\mathrm{high}\big)$.
The term $\prod_{t=1}^{|o_i|}\rho_{i,t}(\theta)$ also appears in the 
objective of \Ours (Equation~\ref{eq:gmpo}). However, instead of applying 
clipping at the sequence-level as in DeepSeek-R1, we find it more 
effective to perform clipping at the token-level, as shown in 
Equation~\ref{eq:gmpo_reward}. The rationale is as follows: (1) Clipping at the token-level is more stable than at the sequence-level. As shown in Figure~\ref{fig:ratio_with_different_clip}, the sequence-level clip (GMPO-seq-clip-$(e^{-0.4}, e^{0.4})$) has a larger importance sampling range than the token-level clip (\Ours$(e^{-0.4}, e^{0.4})$), which makes it more prone to create extreme gradients during optimization. (2) Sequence-level clipping is too aggressive compared to token-level clipping. Once triggered, it sets the gradients of all tokens in the sequence to zero, potentially discarding valuable update signals from informative parts of rollouts.

\noindent\textbf{(ii) Clipping wider.} As illustrated in DAPO~\citep{yu2025dapo}, the clipping operation can limit exploration and cause early deterministic policy, which hinders the scaling process. To encourage exploration without sacrificing stability, DAPO uses a clip-higher strategy, which slightly expands the clipping range $(\epsilon_\mathrm{low}, \epsilon_\mathrm{high})$ from $(0.8, 1.2)$ to $(0.8, 1.28)$. 
As shown in Figure~\ref{fig:1}, we visualize the maximum and minimum importance sampling ratios at each training step for both GRPO and \Ours. The key observations are: (1) As training proceeds, the importance sampling ratio spans a wider range, indicating more aggressive policy updates and increased instability. (2) Compared to GRPO, \Ours preserves a narrower range of importance sampling ratio, suggesting more stable updates. (3) For \Ours, expanding the clipping range from $(e^{-0.2}, e^{0.2})$ to $(-\infty, +\infty)$ increases instability in policy updates.
Based on these findings, we balance training stability with exploration by setting clipping thresholds $(\epsilon_\mathrm{low}, \epsilon_\mathrm{high})$ in Equation~\ref{eq:gmpo} to $(e^{-0.4}, e^{0.4})$. This range is significantly larger than both GRPO and DAPO, encouraging greater exploration and improving performance.

\section{Experiment}
\label{sec:experiment}

\subsection{Implementation Detail}

\noindent\textbf{Model.} We evaluate the algorithm's performance on both language-only and multimodal reasoning tasks.
For the language-only task, following Dr.GRPO~\citep{drgrpo}, we use Qwen2.5-Math-1.5B~\citep{yang2024qwen2}, Qwen2.5-Math-7B, DeepSeek-R1-Distill-Qwen-7B~\citep{guo2025deepseek} and Qwen3-32B~\citep{yang2025qwen3} as our base models to assess performance on mathematical tasks.
For the multimodal task, we use Qwen2.5-VL-Instruct-7B~\citep{bai2025qwen2} as the base model to train GRPO and \Ours, and evaluate their performance on geometry reasoning tasks.


\noindent\textbf{Training.} For the language-only task, following the setup of Dr.GRPO~\citep{drgrpo}, we use MATH~\citep{hendrycks2021measuring} Levels 3–5 as the training dataset for models under 7B, which contains 8,523 mathematical problems. For each question, we generate 8 rollouts and cap the model’s maximum response length at 3,000 tokens.
During each RL training round, the old policy $\pi_{\theta_{\mathrm{old}}}$ produces 1,024 rollouts, and the current policy $\pi_{\theta}$ is updated 8 times with a batch size of 128.
For the Mixture-of-Experts models ($e.g.$, Table~\ref{tab:rl_mm_performance}), we use DeepScaleR~\citep{deepscaler2025} and CountDown~\citep{countdown2025dataset} as the training dataset, with further details provided in Appendix~\ref{app:moe}.
For the multimodal task, we follow the setup of EasyR1~\citep{zheng2025easyr1} and use Geometry3K~\citep{geometry3k} as the training dataset.
All models under 7B are trained on a server with 8$\times$A800 GPUs.
For mathematical problems, rewards are verifiable: ``$1$'' for correct responses and ``$0$'' for incorrect ones. Our method is mainly compared with Dr.GRPO and GRPO, under the same experimental setup as in Tables~\ref{tab:rl_performance_grpo}, \ref{tab:rl_mm_performance}, and \ref{tab:rl_performance}.

\begin{table*}[t]
    \setlength{\tabcolsep}{4.5pt}
    \footnotesize
    \centering
    \caption{Comparison of GRPO and GMPO on five mathematical reasoning benchmarks.}
    \begin{tabular}{l|ccccc|c}
\toprule
Model & AIME24 & AMC & MATH500 & Minerva & Oly. & \textbf{Avg.}\tabularnewline
\midrule
GRPO-1.5B~\cite{shao2024deepseekmath} & 23.3 & 49.4 & 75.2 & 25.7 & 39.0 & 42.5\tabularnewline
\rowcolor{gray0}\Ours-1.5B (Ours) & 20.0 & 53.0 & 77.6 & 30.1 & 38.7 & \textbf{43.9}\tabularnewline

\midrule
GRPO-7B~\cite{shao2024deepseekmath} & 40.0 & 59.0 & 83.4 & 32.4 & 41.3 & 51.2\tabularnewline
\rowcolor{gray0}\Ours-7B (Ours) & 43.3 & 61.4 & 82.0 & 33.5 & 43.6 & \textbf{52.7}\tabularnewline

\midrule
GRPO-7B~\cite{shao2024deepseekmath}  (R1-Distill) & 43.3  & 67.5 & 89.0 & 39.7 & 56.7 & 59.3 \tabularnewline
\rowcolor{gray0}\Ours-7B (R1-Distill, Ours) & 46.6 & 78.3 & 91.4 & 37.9 & 62.5 & \textbf{63.4}\tabularnewline
\bottomrule
\end{tabular}

    \label{tab:rl_performance_grpo}
\end{table*}

\begin{table*}[t]
\centering
\caption{Comparison of GRPO and GMPO for multimodal models (left) and Mixture-of-Experts models (right).}
\begin{subtable}{0.48\linewidth}
\centering
\begin{tabular}{lc}
\toprule
Multimodal-Model & Geometry3K\\
\midrule
GRPO-7B~\citep{shao2024deepseekmath} & 53.3 \\
\rowcolor{gray0}GMPO-7B (Ours) & \textbf{54.7} \\
\bottomrule
\end{tabular}
\end{subtable}
\hfill
\begin{subtable}{0.48\linewidth}
\centering
\begin{tabular}{lc}
\toprule
MoE-Model & MATH500 \\
\midrule
GRPO-32B~\citep{shao2024deepseekmath} & 94.6 \\
\rowcolor{gray0}GMPO-32B (Ours) & \textbf{96.7} \\
\bottomrule
\end{tabular}
\end{subtable}
\label{tab:rl_mm_performance}
\end{table*}

\begin{table*}[t]
    \setlength{\tabcolsep}{4.5pt}
    \footnotesize
    \centering
    \caption{Comparison of GMPO and state-of-the-art methods on mathematical reasoning benchmarks.}
    \begin{tabular}{l|ccccc|c}
\toprule
Model & AIME24 & AMC & MATH500 & Minerva & Oly. & \textbf{Avg.}\tabularnewline
\midrule
Qwen2.5-Math-1.5B~\cite{qwen2025qwen25technicalreport} & 16.7 & 43.4 & 61.8 & 15.1 & 28.4 & 33.1\tabularnewline
Qwen2.5-Math-1.5B-Instruct~\cite{qwen2025qwen25technicalreport} & 10.0 & 48.2 & 74.2 & 26.5 & 40.2 & 39.8\tabularnewline
Oat-Zero-1.5B~\cite{drgrpo} & 20.0 & 53.0 & 74.2 & 25.7 & 37.6 & 42.1\tabularnewline
\rowcolor{gray0}\Ours-1.5B (Ours) & 20.0 & 53.0 & 77.6 & 30.1 & 38.7 & \textbf{43.9}\tabularnewline

\midrule
Qwen2.5-Math-7B~\cite{qwen2025qwen25technicalreport} & 16.7 & 38.6 & 50.6 & 9.9 & 16.6 & 26.5\tabularnewline
SimpleRL-Zero-7B~\cite{zeng2025simplerl} & 26.7 & 60.2 & 78.2 & 27.6 & 40.3 & 46.6\tabularnewline
PRIME-Zero-7B~\cite{prime} & 16.7 & 62.7 & 83.8 & 36.0 & 40.9 & 48.0\tabularnewline
OpenReasoner-Zero-7B @ 3k~\cite{open_reasoner} & 13.3 & 47.0 & 79.2 & 31.6 & 44.0 & 43.0\tabularnewline
OpenReasoner-Zero-7B @ 8k~\cite{open_reasoner} & 13.3 & 54.2 & 82.4 & 31.6 & 47.9 & 45.9\tabularnewline
Eurus-7B~\cite{eurus} & 16.7 & 62.7 & 83.8 & 36.0 & 40.9 & 48.0\tabularnewline
GPG-7B~\cite{chu2025gpg} & 33.3 & 65.0 & 80.0 & 34.2 & 42.4 & 51.0\tabularnewline
Oat-Zero-7B~\cite{drgrpo} & 43.3 & 62.7 & 80.0 & 30.1 & 41.0 & 51.4\tabularnewline
\rowcolor{gray0}\Ours-7B (Ours) & 43.3 & 61.4 & 82.0 & 33.5 & 43.6 & \textbf{52.7}\tabularnewline

\midrule
Oat-Zero-7B~\cite{drgrpo} (R1-Distill) & 50.0 & 74.7 & 89.6 & 37.5 & 55.7 & 61.5 \tabularnewline
\rowcolor{gray0}\Ours-7B (R1-Distill, Ours) & 46.6 & 78.3 & 91.4 & 37.9 & 62.5 & \textbf{63.4}\tabularnewline
\bottomrule
\end{tabular}

    \label{tab:rl_performance}
\end{table*}

\noindent\textbf{Evaluation.} We evaluate our method on five mathematical reasoning benchmarks of varying difficulty following Dr.GRPO~\citep{drgrpo} and one multimodal reasoning benchmark following EasyR1~\citep{zheng2025easyr1}: AIME24, which consists of 30 high-school level olympiad problems from the American Invitational Mathematics Examination 2024; 
AMC, containing 83 intermediate-difficulty multiple-choice problems; 
MATH500, a subset of 500 problems from the original MATH dataset covering algebra, geometry, and number theory; 
Minerva~\citep{lewkowycz2022solving}, featuring 272 graduate-level problems requiring multi-step reasoning; 
and OlympiadBench (Oly.)~\citep{olympiadbench}, a collection of 675 high-difficulty olympiad problems. These benchmarks collectively cover a broad spectrum of problem types and difficulty levels. Geometry3K~\citep{geometry3k} is a visual question answering dataset that consists of a set of 601 questions focused on geometric problem-solving.
We primarily use the Pass@1 metric for comparative analysis. This metric evaluates whether a single generated response to a problem meets the required criteria. For language tasks, we set the temperature to 0.0 and generate one answer per question, following Dr.GRPO~\citep{drgrpo}. For the multimodal task, we set the temperature to 0.5 and generate 16 answers for each question.

\subsection{Performance}
Table~\ref{tab:rl_performance_grpo}~\ref{tab:rl_mm_performance}~\ref{tab:rl_performance} present comprehensive evaluation of our \Ours approach against established reasoning methods across multiple benchmarks. Our method demonstrates consistent and substantial improvements over strong baseline systems.

\noindent{\textbf{Language-only task.}} 
\Ours demonstrates consistent improvements across different base models. With Qwen2.5-Math-1.5B, it achieves 43.9\% average performance, outperforming GRPO by 1.4\% and Dr.GRPO by 1.8\%. Similar gains are observed with Qwen2.5-Math-7B (+1.5\% vs. GRPO, +1.3\% vs. Dr.GRPO) and DeepSeek-R1-Distill-Qwen-7B (+4.1\% vs. GRPO, +1.9\% vs. Dr.GRPO).
In the stability-sensitive Mixture-of-Experts (MoE) setting with Qwen3-32B, GMPO achieves 96.7\% accuracy on MATH500, outperforming GRPO by 2.1\%. Additional results for MoE models are provided in Appendix~\ref{app:moe}.

\noindent{\textbf{Multimodal task.}} Using Qwen2.5-VL-Instruct-7B as the base model, \Ours surpasses GRPO by 1.4\% on Geometry3K, highlighting its potential for broader application in multimodal tasks.

\subsection{Ablation Studies}
\begin{figure}[t]
    \centering
    \includegraphics[width=1.0\linewidth]{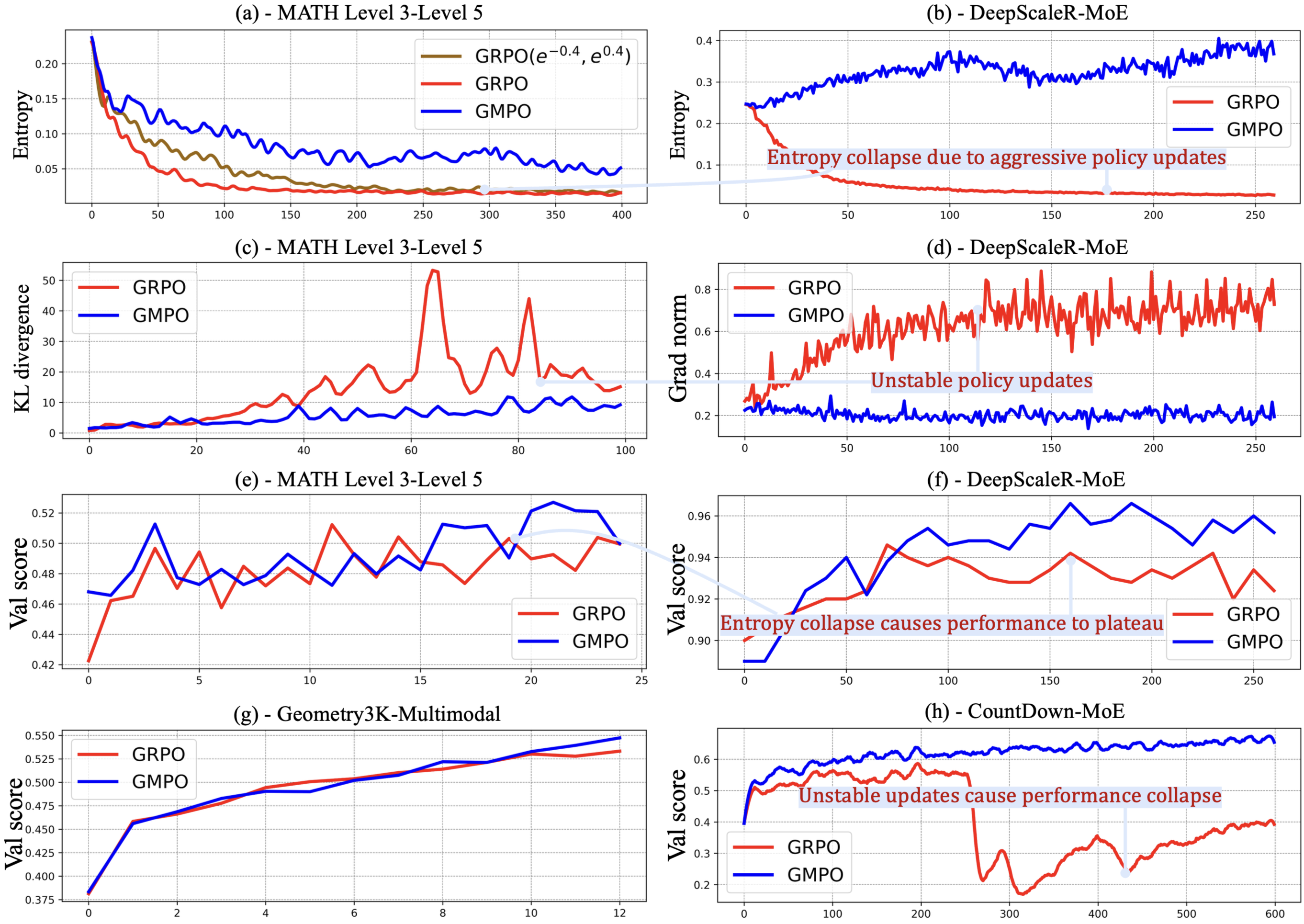}
    \caption{Analysis of entropy, KL divergence, gradient norm, validation score over training steps. (a–b) \Ours maintains higher entropy than GRPO, whether trained on MATH Level 3–Level 5 or DeepScaleR dataset.
    (c-d) \Ours maintains more stable gradient and a smaller KL divergence from the pre-RL model than GRPO. 
    (e–h) \Ours outperforms GRPO in validation scores across language-only and multimodal tasks, for both dense and Mixture-of-Experts models.}
    \label{fig:gmpo_kl_divergence}
\end{figure}

Table~\ref{tab:rl_ablation} presents an ablation study of the key modifications in \Ours relative to GRPO. The effect of the clipping thresholds is presented in Table~\ref{tab:rl_clip_range_term}, and training statistics are shown in Figure~\ref{fig:gmpo_kl_divergence}.

\noindent\textbf{Geometric mean $vs.$ Arithmetic mean.} The performance of GRPO and \Ours is reported in lines $\mathit{1}$ and $\mathit{5}$ of Table~\ref{tab:rl_ablation}, respectively. GRPO achieves an average performance of 51.2\% by optimizing the arithmetic mean of token-level rewards. In contrast, \Ours improves this to 52.7\%, outperforming GRPO by 1.5\%, by optimizing the geometric mean instead. 
In row $\textit{4}$ of Table~\ref{tab:rl_ablation}, we test removing the normalization term ``$\frac{1}{|o|}$'' from the training objective, similar to Dr. GRPO~\citep{drgrpo}. This results in a 0.7\% drop in average performance (52.0\% $vs.$ 52.7\%), suggesting that the normalization term is crucial for maintaining optimal performance.

\begin{table*}[tbp]
    \centering
    \caption{Comparison of objectives and their performance under same training settings.}
\begin{tabular}{l|ccccc|c}
\toprule 

\multicolumn{7}{l}{

$
\textit{1}:
\frac{1}{|o|}\sum_{t=1}^{|o|} \Big(\min \big [\rho_{t}(\theta)\hat{A}, \mathrm{clip}(\rho_{t}(\theta), \epsilon_\mathrm{low}, \epsilon_\mathrm{high}) \hat{A}\big]\Big)
$
 
 }\tabularnewline

\multicolumn{7}{l}{
$
\textit{2}:
\Big\{\prod_{t=1}^{|o|}\big|\rho_{t}(\theta)\hat{A}\big|\Big\}^{\frac{1}{|o|}}\cdot\mathrm{sgn}(\hat{A})
$
}\tabularnewline

\multicolumn{7}{l}{

$
\textit{3}:
\Big\{\big|\min \big [(\prod_{t=1}^{|o|}\rho_{t}(\theta))\hat{A}, \mathrm{clip}(\prod_{t=1}^{|o|}\rho_{t}(\theta), \epsilon_\mathrm{low}, \epsilon_\mathrm{high}) \hat{A}\big ]\big|\Big\}^{\frac{1}{|o|}}\cdot\mathrm{sgn}(\hat{A})
$
 
 }\tabularnewline
\multicolumn{7}{l}{

$
\textit{4}:
\Big\{\prod_{t=1}^{|o|}\big|\min \big [\rho_{t}(\theta)\hat{A}, \mathrm{clip}(\rho_{t}(\theta), \epsilon_\mathrm{low}, \epsilon_\mathrm{high}) \hat{A}\big ]\big|\Big\}\cdot\mathrm{sgn}(\hat{A})
$
 
 }\tabularnewline
\multicolumn{7}{l}{

$
\textit{5}:
\Big\{\prod_{t=1}^{|o|}\big|\min \big [\rho_{t}(\theta)\hat{A}, \mathrm{clip}(\rho_{t}(\theta), \epsilon_\mathrm{low}, \epsilon_\mathrm{high}) \hat{A}\big ]\big|\Big\}^{\frac{1}{|o|}}\cdot\mathrm{sgn}(\hat{A})
$

}\tabularnewline
\midrule
Training objectives & AIME24 & AMC & MATH500 & Minerva & Oly. & \textbf{Avg.}\tabularnewline
\midrule
\textit{0} (Pre-RL model) & 16.7 & 38.6 & 50.6 & 9.9 & 16.6 & 26.5\tabularnewline
\textit{1} (GRPO) & 40.0 & 59.0 & 83.4 & 32.4 & 41.3 & 51.2\tabularnewline
\textit{2} (\textit{without} clip) & 40.0 & 63.9 & 80.6 & 33.5 & 43.7 & 52.3\tabularnewline
\textit{3} (\textit{with} seq-clip) & 46.6 & 57.8 & 80.2 & 34.2 & 44.3 & 52.6\tabularnewline
\textit{4} (\textit{without} norm)& 36.6 & 67.4 & 82.0 & 29.8 & 44.1 & 52.0\tabularnewline
\textit{5} (GMPO) & 43.3 & 61.4 & 82.0 & 33.5 & 43.6 & \textbf{52.7}\tabularnewline
\bottomrule
\end{tabular}
    \label{tab:rl_ablation}
\end{table*}

\begin{table*}[tbp]
    \small
    \centering
    \caption{Influence of the clipping thresholds on model performance.}
    \begin{tabular}{c|c|ccccc|c}
\toprule
& Clipping thresholds $(\epsilon_\mathrm{low}, \epsilon_\mathrm{high})$ & AIME24 & AMC & MATH500 & Minerva & Oly. & \textbf{Avg.}\tabularnewline
\midrule
\textit{1} & $(e^{-0.2}, e^{0.2})$ & 36.6 & 60.2 & 84.2 & 35.7 & 45.0 & 52.4\tabularnewline
\textit{2} & $(e^{-0.4}, e^{0.4})$ & 43.3 & 61.4 & 82.0 & 33.5 & 43.6 & \textbf{52.7}\tabularnewline
\textit{3} & $(e^{-0.8}, e^{0.8})$ & 40.0 & 60.2 & 82.2 & 33.5 & 44.7 & 52.1\tabularnewline
\textit{4} & $(-\infty,+\infty)$ & 40.0 & 63.9 & 80.6 & 33.5 & 43.7 & 52.3\tabularnewline
\bottomrule
\end{tabular}

    \label{tab:rl_clip_range_term}
\end{table*}

\noindent\textbf{Clipping strategy.} The performance of \Ours without clip, with token-level clip, and with sequence-level clip are shown in lines \textit{2}, \textit{3}, and \textit{5}, respectively. The corresponding ranges of importance sampling ratios are shown in Figure~\ref{fig:ratio_with_different_clip}, labeled as \Ours$(e^{-0.4}, e^{0.4})$, \Ours-seq-clip-$(e^{-0.4}, e^{0.4})$, and GRPO$(0.8,1.2)$. Clipping at the sequence-level achieves similar performance to token-level clipping but has a larger range of importance sampling ratios. Therefore, we use the token-level clipping strategy.
Removing the clipping range term (\Ours$(-\infty,+\infty)$) leads to excessive fluctuations in the importance sampling ratio during training, which affects stability and results in a 0.4\% decrease in average performance (52.3\% vs. 52.7\%).

\noindent\textbf{Influence of the clipping thresholds.} 
To find the optimal clipping thresholds for \Ours, we train the model under different clipping thresholds, as shown in Table~\ref{tab:rl_clip_range_term} and Figure~\ref{fig:ratio_with_different_clip}. A larger clipping range encourages exploration but introduces instability to optimization, which ultimately affects performance. To balance stability and performance, we set $(\epsilon_\mathrm{low}, \epsilon_\mathrm{high})$ in Equation~\ref{eq:gmpo} to $(e^{-0.4}, e^{0.4})$, which has a stable range of importance sampling ratio and achieves the best performance.

\noindent\textbf{Exploration capability.} As noted in~\citep{cui2025entropy_mechanism}, language models in reinforcement learning often trade off entropy for short-term performance. Premature entropy collapse can cause performance to plateau.
As shown in Figure~\ref{fig:gmpo_kl_divergence} (a-b), we visualize the mean token entropy of \Ours and GRPO when training the policy model at MATH Level 3-Level 5 and the more challenging mathematical dataset DeepScaleR.
GRPO’s entropy drops rapidly during training, limiting exploration and causing performance to plateau (Figure~\ref{fig:gmpo_kl_divergence} (e–g)). As shown in Figure~\ref{fig:gmpo_kl_divergence} (a), applying a wider clipping range for GRPO temporarily encourages exploration, but entropy still declines quickly over time. This behavior arises because GRPO optimizes the arithmetic mean of token-level rewards, which is sensitive to outliers. Consequently, it can generate aggressive updates that sharply reduce entropy while offering only marginal performance gains, hindering both exploration and scalability.

In contrast, \Ours employs the geometric mean, which is robust to outliers. This allows it to maintain stable, moderate entropy, enabling consistent exploration throughout training and resulting in higher rewards and better overall performance than GRPO, as shown in Figure~\ref{fig:gmpo_kl_divergence} (e–g).

\noindent\textbf{Training stability.} As shown in Figure~\ref{fig:gmpo_kl_divergence} (c-d), we visualize the gradient norm during training, and the KL divergence between the current model $\pi_{\theta}$ and the reference model $\pi_{\mathrm{ref}}$. $\pi_{\mathrm{ref}}$ is initialized as the base model before RL training. As training progresses, \Ours maintains stable gradient and a low KL divergence from the reference model, indicating greater training stability and a lower risk of overfitting. In contrast, GRPO exhibits unstable gradient and large KL divergence, suggesting unstable learning and a greater tendency to drift away from the reference model. 

\noindent\textbf{Validation scores.} Figure~\ref{fig:gmpo_kl_divergence} (e–h) shows validation scores under different training settings. Figures (e) and (f, g) correspond to Tables~\ref{tab:rl_performance_grpo} and \ref{tab:rl_mm_performance}, respectively, while results on CountDown are detailed in Appendix~\ref{app:moe}. GMPO consistently outperforms GRPO in validation scores across language-only (e, f, h) and multimodal (g) tasks, for both dense (e, g) and Mixture-of-Experts models (f, h).
\section{Conclusion}
\label{sec:conclusion}

{We propose GMPO, a stabilized variant of GRPO. By optimizing the geometric mean of token-level rewards and enlarging the clipping range of importance sampling ratio, \Ours not only alleviates the instability in policy updates but also enhances exploration capabilities,
as evidenced by a narrower objective value range, more stable gradients, and consistently lower KL divergence with higher token entropy throughout training.
Extensive experiments on language-only and multimodal reasoning benchmarks demonstrate that \Ours outperforms GRPO in terms of both stability and reasoning capacity. 
This work sets the stage for future research on developing more reliable and scalable RL systems.}

\clearpage
%
%

\bibliography{main}
\bibliographystyle{iclr2026_conference}

\newpage
\appendix
\begin{center}
\bfseries \Large Appendices
\end{center}

\section{Gradient Derivation}
\label{app:derivation}
To better understand why \Ours is more stable than GRPO, we analyze its robustness to tokens with extreme importance sampling ratios from a gradient perspective. Specifically, we first derive the gradient of the importance sampling ratio $\rho_{i,t}(\theta)$ with respect to the model parameter $\theta$ in \textbf{Lemma 1}. Building on this result, we then derive the gradients of GRPO and GMPO with respect to $\theta$ in \textbf{Lemmas 2} and \textbf{3}. For clarity, the clipping range term is omitted in the gradient derivation.

\textbf{Lemma 1 (Derivative of the importance sampling ratio)} 
\begin{align}
\scalebox{1.2}{$\displaystyle \nabla$}_\theta\rho_{i,t}(\theta) &= \scalebox{1.2}{$\displaystyle \nabla$}_\theta\frac{\pi_{\theta}(o_{i,t}|q,o_{i,<t})}{\pi_{\theta_{\mathrm{old}}}(o_{i,t}|q,o_{i,<t})}\notag\\
&= \frac{1}{\pi_{\theta_{\mathrm{old}}}(o_{i,t}|q,o_{i,<t})}\scalebox{1.2}{$\displaystyle \nabla$}_\theta\pi_{\theta}(o_{i,t}|q,o_{i,<t})\notag\\
&= \frac{\pi_{\theta}(o_{i,t}|q,o_{i,<t})}{\pi_{\theta_{\mathrm{old}}}(o_{i,t}|q,o_{i,<t})}\cdot \frac{1}{\pi_{\theta}(o_{i,t}|q,o_{i,<t})}\cdot \scalebox{1.2}{$\displaystyle \nabla$}_\theta\pi_{\theta}(o_{i,t}|q,o_{i,<t})\notag\\
&= \rho_{i,t}(\theta) \scalebox{1.2}{$\displaystyle \nabla$}_\theta\mathrm{log}(\pi_{\theta}(o_{i,t}|q,o_{i,<t}))\notag
\end{align}

\textbf{Lemma 2 (Derivative of the GRPO objective)} 
\begin{align}
\scalebox{1.2}{$\displaystyle \nabla$}_\theta\mathcal{J}^*_{\mathrm{GRPO}}(\pi_\theta)\Big|_{q, o_i} 
&= \scalebox{1.2}{$\displaystyle \nabla$}_\theta \frac{1}{G}\sum_{i=1}^G\frac{1}{|o_i|}\sum_{t=1}^{|o_i|}\rho_{i,t}(\theta)\hat{A}_i\notag\\
&= 
\frac{1}{G\cdot|o_i|}\sum_{t=1}^{|o_i|}\cdot\hat{A}_i\cdot \scalebox{1.2}{$\displaystyle \nabla$}_\theta\rho_{i,t}(\theta)\notag\\
&= 
\frac{1}{G\cdot|o_i|}\sum_{t=1}^{|o_i|}{\color[rgb]{1.0,0,0}\rho_{i,t}(\theta)}\cdot \hat{A}_i\cdot\scalebox{1.2}{$\displaystyle \nabla$}_\theta \mathrm{log}(\pi_{\theta}(o_{i,t}|q,o_{i,<t}))\notag
\end{align}

\textbf{Lemma 3 (Derivative of the GMPO objective)} 
\begin{align}
\scalebox{1.2}{$\displaystyle \nabla$}_\theta\mathcal{J}^*_{\mathrm{GMPO}}(\pi_\theta)\Big|_{q, o_i} &= \scalebox{1.2}{$\displaystyle \nabla$}_\theta\frac{1}{G}\sum_{i=1}^G\Big(\prod_{t=1}^{|o_i|}\big|\rho_{i,t}(\theta)\hat{A}_i\big|\Big)^{\frac{1}{|o_i|}}\cdot\mathrm{sgn}(\hat{A}_i)\notag\\
&= \scalebox{1.2}{$\displaystyle \nabla$}_\theta\frac{1}{G}\sum_{i=1}^G\Big(\prod_{t=1}^{|o_i|}\rho_{i,t}(\theta)\Big)^{\frac{1}{|o_i|}}\cdot\hat{A}_i\notag\\
&= \frac{1}{G\cdot|o_i|} \Big(\prod_{t=1}^{|o_i|}\rho_{i,t}(\theta)\Big)^{\frac{1}{|o_i|}-1}\cdot\hat{A}_i\cdot\scalebox{1.2}{$\displaystyle \nabla$}_\theta\prod_{t=1}^{|o_i|}\rho_{i,t}(\theta)\notag\\
&= \frac{1}{G\cdot|o_i|} \Big(\prod_{t=1}^{|o_i|}\rho_{i,t}(\theta)\Big)^{\frac{1}{|o_i|}-1}\cdot\hat{A}_i\cdot\sum_{k=1}^{|o_i|}\Big(\prod_{t=1,t\neq k}^{|o_i|}\rho_{i,t}(\theta)\Big)\scalebox{1.2}{$\displaystyle \nabla$}_\theta\rho_{i,k}(\theta)\notag\\
&= \frac{1}{G\cdot|o_i|} \Big(\prod_{t=1}^{|o_i|}\rho_{i,t}(\theta)\Big)^{\frac{1}{|o_i|}-1}\cdot\hat{A}_i\cdot\sum_{k=1}^{|o_i|}\Big(\prod_{t=1}^{|o_i|}\rho_{i,t}(\theta)\Big)\scalebox{1.2}{$\displaystyle \nabla$}_\theta\mathrm{log}(\pi_{\theta}(o_{i,k}|q,o_{i,<k}))\notag\\
&= \frac{1}{G\cdot|o_i|}\sum_{k=1}^{|o_i|}{\color[rgb]{0,0,1.0}\Big(\prod_{t=1}^{|o_i|}\rho_{i,t}(\theta)\Big)^{\frac{1}{|o_i|}}}\cdot \hat{A}_i\cdot\scalebox{1.2}{$\displaystyle \nabla$}_\theta \mathrm{log}(\pi_{\theta}(o_{i,k}|q,o_{i,<k}))\notag
\end{align}

\section{Performance on Mixture-of-Experts Models}
\label{app:moe}
To better demonstrate the stability advantage of GMPO over GRPO, we conduct post-training experiments on Mixture-of-Experts (MoE) models, where stability is particularly critical. The experiments are performed on the DeepScaleR~\citep{deepscaler2025} and CountDown~\citep{countdown2025dataset} datasets, with detailed training settings provided in Table~\ref{tab:rl_moe_settings}.
Specifically, DeepScaleR consists of approximately 40,000 unique mathematics problem-answer pairs compiled from AIME~\citep{li2024numinamath}, AMC~\citep{li2024numinamath}, Omni-MATH dataset, and Still dataset. CountDown consists of arithmetic puzzles where models combine given numbers using basic operations to reach a target, commonly used to test algorithmic reasoning and step-by-step problem solving. We reserve a subset of the CountDown dataset for model evaluation.

\begin{figure}[tbp]
    \centering
    \includegraphics[width=1.0\linewidth]{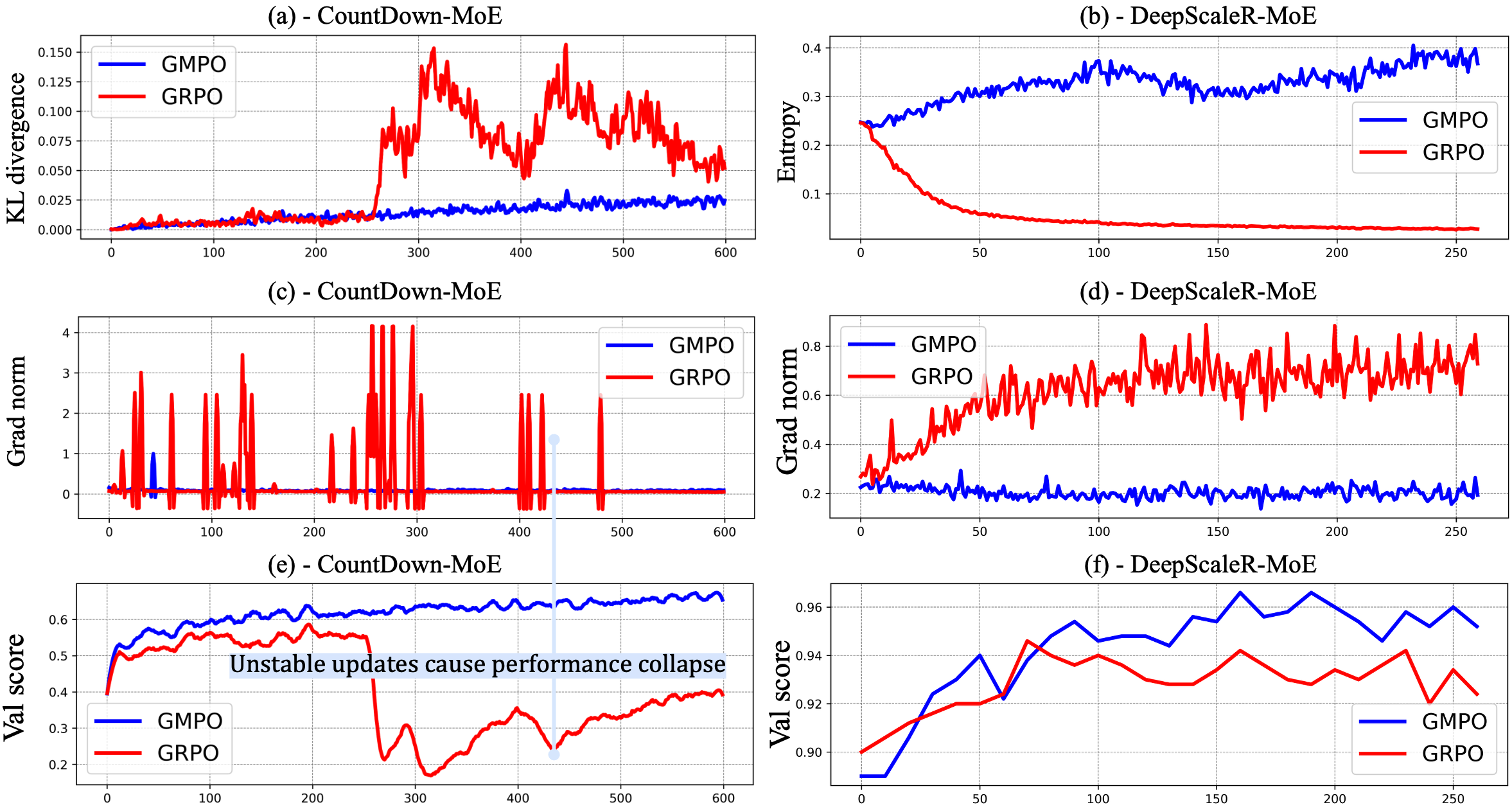}
    \caption{Analysis of entropy, KL divergence, gradient norm, and validation score over training steps on Mixture-of-Experts models. (a) \Ours maintains smaller KL divergence than GRPO.
    (b) \Ours maintains higher entropy than GRPO.
    (c-d) \Ours maintains more stable gradient norm than GRPO, suggesting more stable policy optimization. 
    (e-f) \Ours achieves higher validation score than GRPO.}
    \label{fig:gmpo_moe}
\end{figure}

\noindent\textbf{CountDown.} As shown in Figure~\ref{fig:gmpo_moe}(a)(c), we visualize the KL divergence and gradient norm during GMPO and GRPO training. GMPO consistently maintains a lower KL divergence from the reference model and a steadier gradient norm than GRPO. Consequently, GMPO achieves stable validation scores, whereas GRPO collapses after about 250 steps, as shown in Figure~\ref{fig:gmpo_moe} (e).

\noindent\textbf{DeepScaleR.} As shown in Figure~\ref{fig:gmpo_moe} (b)(d), GMPO achieves higher entropy while a steadier gradient norm than GRPO. Consequently, GMPO achieves higher validation scores as shown in Figure~\ref{fig:gmpo_moe} (f).

\begin{table*}[tbp]
    \centering
    \small
    \caption{Training settings for GMPO and GRPO on Mixture-of-Experts models. Qwen2.5‑200M$^{\dagger}$ is a small-scale language model adapted from the Qwen2.5 series~\citep{qwen2025qwen25technicalreport}. “Bs / Mini Bs” denote the batch size and mini-batch size used during training, respectively. “E./Act. E.” indicate the total number of experts in the model and the number of experts activated per token, respectively.}
    \begin{tabular}{ccccc}
\toprule
Training dataset & Eval dataset & Base model & Bs./Mini Bs. & E./Act. E.\tabularnewline
\midrule
DeepScaleR & MATH500 & Qwen3-32B~\cite{yang2025qwen3} & 128/64 & 128/8\tabularnewline
CountDown & CountDown(Val) & Qwen2.5-200M$^{\dagger}$~\cite{bai2025qwen2} & 256/128 & 8/1\tabularnewline
\bottomrule
\end{tabular}

    \label{tab:rl_moe_settings}
\end{table*}

\section{Analysis of the Normalization Factor in the Geometric-Mean}
\begin{figure}[tbp]
    \centering
    \includegraphics[width=0.7\linewidth]{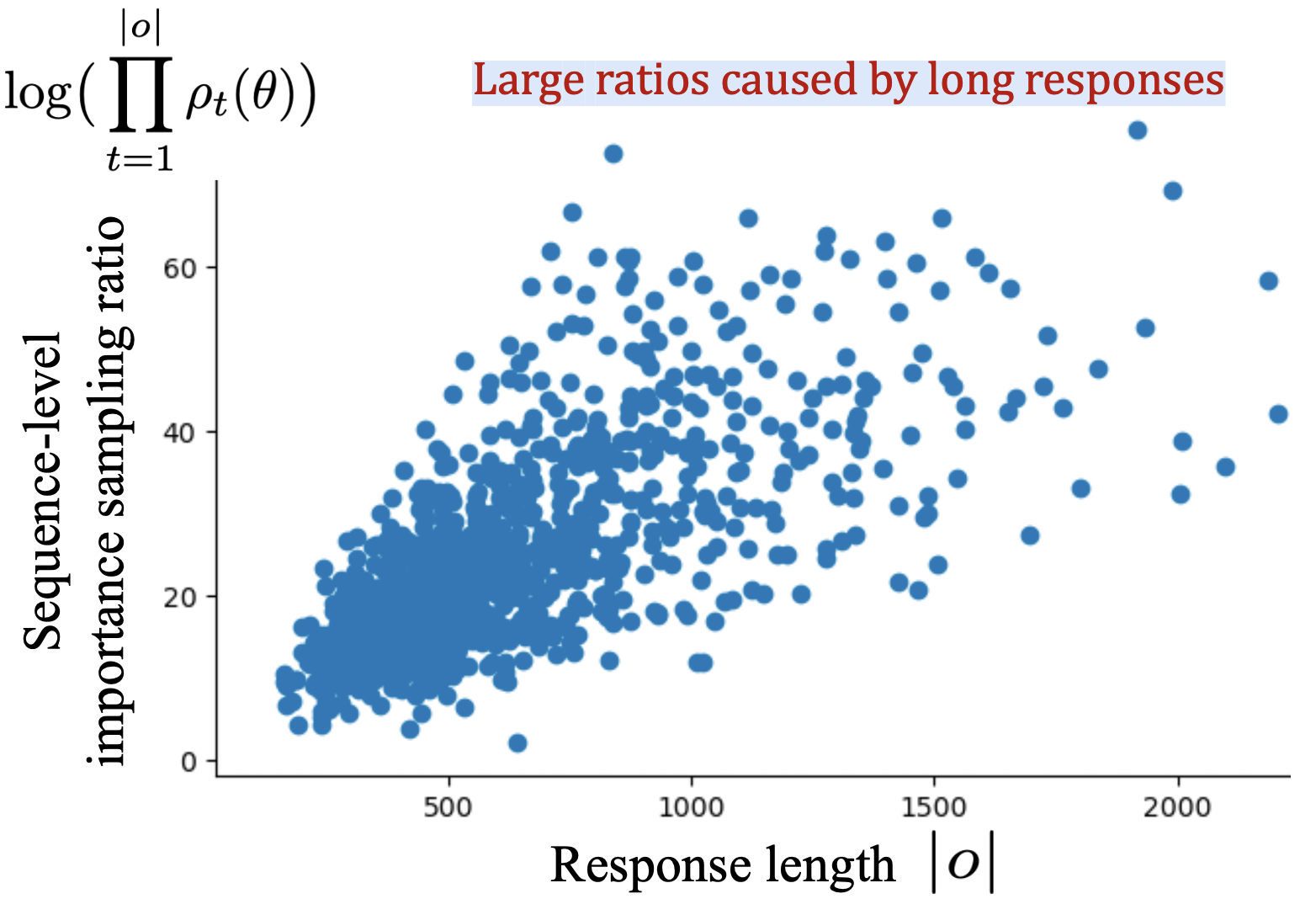}
    \caption{Sequence-level importance sampling ratios from trajectories that yield positive rewards during GRPO training. Without normalization, these ratios can become highly unstable, especially as the response length increases.}
    \label{fig:gmpo_seq_ratio}
\end{figure}

Unlike vanilla GRPO in DeepSeek-math~\citep{shao2024deepseekmath}, 
DeepSeek-R1~\citep{deepseekr1} maximizes the sequence-level reward $(\prod_{t=1}^{|o_i|}\rho_{i,t}(\theta))\hat{A}_i$.
The term $\prod_{t=1}^{|o_i|}\rho_{i,t}(\theta)$ also appears in the objective of \Ours (Equation~\ref{eq:gmpo}). 
Unlike DeepSeek-R1, GMPO introduces an additional power-based normalization term: ``$\frac{1}{|o_i|}$'', which we find is critical for GMPO objective.
As shown in Figure~\ref{fig:gmpo_seq_ratio}, we visualize the range of sequence-level importance sampling ratios from trajectories that yield positive rewards during GRPO training. Without the normalization term, these sequence-level importance sampling ratios can become very large, especially as the response length increases.
This phenomenon ultimately leads to unstable policy optimization, which in turn degrades the model’s final performance.

\end{document}